\documentclass[a4paper,twoside]{./template/scitepress/article}

\usepackage{cite}

\usepackage[pdftex]{graphicx}
\graphicspath{{./figures/}}
\DeclareGraphicsExtensions{.pdf,.jpeg,.png}

\usepackage[hidelinks]{hyperref}

\usepackage{pgfplots}
\usepgfplotslibrary{fillbetween,patchplots,groupplots}
\usetikzlibrary{arrows.meta,backgrounds,fit,shapes.geometric,patterns,positioning,matrix, external}
\pgfplotsset{compat=newest}
\tikzexternalize[prefix=figures/]

\tikzset{every picture/.style={/utils/exec={\small\sffamily}}}

\pgfmathsetseed{42}

\pgfmathdeclarefunction{invgauss}{4}{%
  \pgfmathparse{(sqrt(-2*ln(#1))*cos(deg(2*pi*#2))*#4) + #3}%
}

\pgfmathdeclarefunction{circle_thresh}{3}{%
	\pgfmathparse{ifthenelse(sin(acos(#1 / 6.0)) < abs(#2) / #3, #2, -inf)}%
}

\usepackage{tcolorbox}  %

\usepackage{rotating}

\definecolor{rwth}   {RGB}{  0  84 159}
\definecolor{rwth-75}{RGB}{ 64 127 183}
\definecolor{rwth-50}{RGB}{142 186 229}
\definecolor{rwth-25}{RGB}{199 221 242}
\definecolor{rwth-10}{RGB}{232 241 250}

\definecolor{black}   {RGB}{  0   0   0}
\definecolor{black-75}{RGB}{100 101 103}
\definecolor{black-50}{RGB}{156 158 159}
\definecolor{black-25}{RGB}{207 209 210}
\definecolor{black-10}{RGB}{236 237 237}

\definecolor{magenta}   {RGB}{227   0 102}
\definecolor{magenta-75}{RGB}{233  96 136}
\definecolor{magenta-50}{RGB}{241 158 177}
\definecolor{magenta-25}{RGB}{249 210 218}
\definecolor{magenta-10}{RGB}{253 238 240}

\definecolor{yellow}   {RGB}{255 237   0}
\definecolor{yellow-75}{RGB}{255 240  85}
\definecolor{yellow-50}{RGB}{255 245 155}
\definecolor{yellow-25}{RGB}{255 250 209}
\definecolor{yellow-10}{RGB}{255 253 238}

\definecolor{petrol}   {RGB}{  0  97 101}
\definecolor{petrol-75}{RGB}{ 45 127 131}
\definecolor{petrol-50}{RGB}{125 164 167}
\definecolor{petrol-25}{RGB}{191 208 209}
\definecolor{petrol-10}{RGB}{230 236 236}

\definecolor{turkis}   {RGB}{  0 152 161}
\definecolor{turkis-75}{RGB}{  0 177 183}
\definecolor{turkis-50}{RGB}{137 204 207}
\definecolor{turkis-25}{RGB}{202 231 231}
\definecolor{turkis-10}{RGB}{235 246 246}

\definecolor{grun}   {RGB}{ 87 171  39}
\definecolor{grun-75}{RGB}{141 192  96}
\definecolor{grun-50}{RGB}{184 214 152}
\definecolor{grun-25}{RGB}{221 235 206}
\definecolor{grun-10}{RGB}{242 247 236}

\definecolor{maigrun}   {RGB}{189 205   0}
\definecolor{maigrun-75}{RGB}{208 217  92}
\definecolor{maigrun-50}{RGB}{224 230 154}
\definecolor{maigrun-25}{RGB}{240 243 208}
\definecolor{maigrun-10}{RGB}{249 250 237}

\definecolor{orange}   {RGB}{246 168   0}
\definecolor{orange-75}{RGB}{250 190  80}
\definecolor{orange-50}{RGB}{253 212 143}
\definecolor{orange-25}{RGB}{254 234 201}
\definecolor{orange-10}{RGB}{255 247 234}

\definecolor{rot}   {RGB}{204   7  30}
\definecolor{rot-75}{RGB}{216  92  65}
\definecolor{rot-50}{RGB}{230 150 121}
\definecolor{rot-25}{RGB}{243 205 187}
\definecolor{rot-10}{RGB}{250 235 227}

\definecolor{bordeaux}   {RGB}{161  16  53}
\definecolor{bordeaux-75}{RGB}{182  82  86}
\definecolor{bordeaux-50}{RGB}{205 139 135}
\definecolor{bordeaux-25}{RGB}{229 197 192}
\definecolor{bordeaux-10}{RGB}{245 232 229}

\definecolor{violett}   {RGB}{ 97  33  88}
\definecolor{violett-75}{RGB}{131  78 117}
\definecolor{violett-50}{RGB}{168 133 158}
\definecolor{violett-25}{RGB}{210 192 205}
\definecolor{violett-10}{RGB}{237 229 234}

\definecolor{lila}   {RGB}{122 111 172}
\definecolor{lila-75}{RGB}{155 145 193}
\definecolor{lila-50}{RGB}{188 181 215}
\definecolor{lila-25}{RGB}{222 218 235}
\definecolor{lila-10}{RGB}{242 240 247}

\usepackage{amsmath}
\usepackage{amssymb}
\usepackage{amstext}
\usepackage{amsthm}
\interdisplaylinepenalty=2500

\usepackage{bm}  %
\usepackage{mathtools}

\DeclarePairedDelimiter{\abs}{\lvert}{\rvert}%
\DeclarePairedDelimiter{\norm}{\lVert}{\rVert}%
\DeclarePairedDelimiterX{\infdivx}[2]{(}{)}{%
	#1\;\delimsize\|\;#2%
}
\renewcommand{\vec}[1]{\boldsymbol{\mathbf{#1}}}

\usepackage{array}

\usepackage{siunitx} %

\sisetup{detect-all} %
\sisetup{round-precision=1, round-mode=places}

\usepackage{booktabs}
\usepackage{multirow}

\usepackage{subcaption}

\usepackage[acronym,toc,xindy]{glossaries}

\usepackage{xparse}
\DeclareDocumentCommand{\newdualentry}{ O{} O{} m m m m } {
	\newglossaryentry{gls-#3}{name={#5},text={#5\glsadd{#3}},
		description={#6},#1
	}
	\makeglossaries
	\newacronym[see={[Glossary:]{gls-#3}},#2]{#3}{#4}{#5\glsadd{gls-#3}}
}

\setglossarystyle{list} %
\newacronym{ai}{AI}{artificial intelligence}
\newacronym{ad}{AD}{anomaly detection}
\newacronym{as}{AS}{anomaly segmentation}
\newacronym{dl}{DL}{deep learning}
\newacronym{ml}{ML}{machine learning}
\newacronym{ood}{OOD}{out-of-distribution}
\newacronym{qc}{QC}{quality control}
\newacronym{kd}{KD}{knowledge distillation}
\newacronym{rl}{RL}{reinforcement learning}
\newacronym{gan}{GAN}{generative adversarial network}
\newacronym{ae}{AE}{autoencoder}
\newacronym{vae}{VAE}{variational autoencoder}
\newacronym{aae}{AAE}{adversarial autoencoder}
\newacronym{gmm}{GMM}{Gaussian mixture model}
\newacronym{ccd}{CCD}{charge-coupled device}
\newacronym{roi}{ROI}{region of interest}
\newacronym{svm}{SVM}{support vector machine}
\newacronym{smo}{SMO}{sequential minimal optimization}
\newacronym{fpn}{FPN}{feature pyramid networks}

\newacronym{roc}{ROC}{receiver operating characteristic}
\newacronym{auroc}{AUROC}{area under the \gls{roc} curve}
\newacronym{aupr}{AUPR}{area under the precision-recall curve}
\newacronym{auprosavg}{AUPR\textsubscript{os,avg}}{area under the class-averaged, open set precision-recall curve} %
\newacronym{aupros}{AUPR\textsubscript{os}}{area under the open set precision-recall curve}
\newacronym{iou}{IoU}{intersection over union}
\newacronym{tp}{TP}{true positive}
\newacronym{tn}{TN}{true negative}
\newacronym{fp}{FP}{false positive}
\newacronym{fn}{FN}{false negative}
\newacronym{tpr}{TPR}{true positive rate}
\newacronym{tnr}{TNR}{true negative rate}
\newacronym{fnr}{FNR}{false negative rate}
\newacronym{fpr}{FPR}{false positive rate}
\newacronym{pro}{PRO}{per-region-overlap}
\newacronym{flop}{FLOP}{floating-point operation}
\newacronym{fps}{FPS}{frames per second} %
\newacronym{cca}{CCA}{connected component analysis} %

\newacronym{pdf}{PDF}{probability density function}
\newacronym{cdf}{CDF}{cumulative distribution function}

\newacronym{mse}{MSE}{mean squared error}

\newacronym{cnn}{CNN}{convolutional neural network}
\newacronym{pca}{PCA}{principal component analysis}  %
\newacronym{fc}{FC}{fully connected}
\newacronym{relu}{ReLU}{rectified linear unit}
\newacronym{serlu}{SERLU}{scaled exponentially-regularized linear unit}
\newacronym{oas}{OAS}{oracle approximating shrinkage}  %
\newacronym{npca}{NPCA}{negated principal component analysis}
\newacronym{ann}{ANN}{artificial neural network}
\newacronym{mlp}{MLP}{multi-layer perceptron}
\newacronym{nlp}{NLP}{natural language processing}

\newacronym{gda}{GDA}{Gaussian discriminant analysis}
\newacronym{1nn}{1-NN}{1-nearest neighbor}
\newacronym{ocsvm}{OC-SVM}{one-class \gls{svm}}
\newacronym{svdd}{SVDD}{support vector data description}
\newacronym{if}{IF}{isolation forest}
\newacronym{lof}{LOF}{local outlier factor}
\newacronym{odin}{ODIN}{outlier detection using in-degree number}
\newacronym{osr}{OSR}{open set recognition}
\newacronym[longplural={known known classes}]{kkc}{KKC}{known known class}
\newacronym[longplural={known unknown classes}]{kuc}{KUC}{known unknown class}
\newacronym[longplural={unknown unknown classes}]{uuc}{UUC}{unknown unknown class}
\newacronym[longplural={unknown known classes}]{ukc}{UKC}{unknown unknown class}
\newacronym{msp}{MSP}{maximum softmax probability}
\newacronym{oe}{OE}{outlier exposure}
\newacronym{vrm}{VRM}{vicinal risk minimization}
\newacronym{ebs}{EBS}{energy-based score}
\newacronym{nll}{NLL}{negative log-likelihood}
\newacronym{em}{EM}{expectation maximization}
\newacronym{bic}{BIC}{Bayesian information criterion}
\newacronym{sem}{SEM}{scanning electron microscopy}
\newacronym{serr}{SEM}{standard error of the mean}
\newacronym{maxlogit}{MaxLogit}{maximum of the unnormalized predictions} %
\newacronym{fssd}{FSSD}{feature space singularity distance}
\newacronym{maha}{Maha}{Mahalanobis}
\newacronym{db}{d\&b}{dyed \& bleached}
\newacronym{avi}{AVI}{automated visual inspection}
\newacronym{avis}{AVIS}{\gls{avi} system}
\newacronym{aif}{AiF}{German  Federation  of Industrial  Research  Associations}
\newacronym{en}{EN}{EfficientNet}
\newacronym{ci}{CI}{confidence interval}
\newacronym{ewc}{EWC}{elastic weight consolidation}
\newacronym{nf}{NF}{normalizing flows}
\newacronym{vit}{ViT}{vision transformer}
\newacronym{dpi}{DPI}{dots per inch}
\newacronym{bmbf}{BMBF}{German Federal Ministry of Education and Research}
\newacronym{cagr}{CAGR}{compound annual growth rate}
\newacronym{oscr}{OSCR}{open set class-wise recall}
\newacronym{pet}{PET}{polyethylene terephthalate}
\newacronym{olp}{OLP}{OnLoomPattern}
\newacronym{mida}{MIDA}{Multiple Image Defect Analysis}

\newdualentry{tsne}{t-SNE}{t-distributed stochastic neighbor embedding}{
	A non-linear visualization technique for high dimensional data introduced in \cite{Maaten2008Visualizingdatausing}.
	The algorithm optimizes similarity between the probabilities of picking nearby points in the low-dimensional visualization with a probability distribution modeled in the high-dimensional data
}

\newdualentry{2afc}{2AFC}{two alternative forced choice}{
	A method for measuring the subjective experience of a person through their pattern of choices.
	The subject is presented with two alternative options and is forced to choose one
}

\newdualentry{clahe}{CLAHE}{contrast limited adaptive histogram equalization}{
	An image processing technique to increase contrast.
	Histogram equalization is performed on a local neighborhood of each pixel (by transforming each with the local \gls{cdf}).
	To limit amplification of noise in near-constant regions, the histogram is clipped before computing the \gls{cdf}
}

\newdualentry{loo}{LOO}{leave-one-out}{
	A cross-validation method.
}

\newdualentry{ssim}{SSIM}{structural similarity}{
	A similarity measure between two images for estimating perceived quality.
	The measure is calculated on $11\times 11$ windows in the images $x$ and $y$ as
		\begin{equation*}
		\operatorname{SSIM}(x, y) = \frac{\left(2\mu_x\mu_y + C_1\right)\left(2\sigma_{xy} + C_2\right)}{\left(\mu_x^2 + \mu_y^2 + C_1\right)\left(\sigma_x^2 + \sigma_y^2 + C_2\right)},
		\end{equation*}
	where $\mu_x, \mu_y$ are the averages of $x$ and $y$, $\sigma_x^2, \sigma_y^2$ their variances, $\sigma_{xy}$ is the covariance of $x$ and $y$ and $C_1, C_2$ are constants \cite{Wang2004Imagequalityassessment}
}

\newdualentry{sgd}{SGD}{stochastic gradient descent}{
	Popular algorithm by which deep learning algorithms are trained.
}

\newdualentry{bce}{BCE}{binary crossentropy}{
	The \gls{bce} is defined as \begin{equation*}
		\operatorname{\gls{bce}}\left(\Phi(\vec{x}; \mathcal{W}), y\right) = - y\log(\Phi(\vec{x}; \mathcal{W})) - (1 - y) \log(1 - \Phi(\vec{x}; \mathcal{W})), %
	\end{equation*}
	where $\Phi$ denotes a neural network parametrized by $\mathcal{W}$ applied to an image $\vec{x}$, and $y$ denotes whether an image is considered normal ($y = 0$) or anomalous ($y = 1$). \cite{Hastie2009ElementsStatisticalLearning}
}

\newdualentry{hsc}{HSC}{hypersphere classifier}{
	The \gls{hsc} is defined as \begin{equation*} %
		\begin{split}
			\operatorname{\gls{hsc}}\left(\Phi(\vec{x}; \mathcal{W}), y\right) = &(1 - y)\norm{\Phi(\vec{x}; \mathcal{W})}^2 \\
			& - y \log\left(1 - \exp(-\norm{\Phi(\vec{x}; \mathcal{W})}^2\right).
		\end{split}
	\end{equation*}
	and was proposed in \cite{Ruff2020RethinkingAssumptionsDeep} to enforce the clustering of normal data in latent representations.
}

\newdualentry{fl}{FL}{focal loss}{
	The \gls{fl} is defined as \begin{equation*}
		\begin{split}
			\operatorname{\gls{fl}}\left(\Phi(\vec{x}; \mathcal{W}), y\right) = &- y (1 - \Phi(\vec{x}; \mathcal{W}))^{\gamma}\log(\Phi(\vec{x}; \mathcal{W})) \\
			& - (1 - y) \Phi(\vec{x}; \mathcal{W})^{\gamma} \log(1 - \Phi(\vec{x}; \mathcal{W})), %
		\end{split}
	\end{equation*}
	with $\gamma$ being the focussing parameter that can be used to put increasing focus on misclassified examples.
	\Gls{fl} was proposed in \cite{Lin2020FocalLossDense}.
}

\newdualentry{spade}{SPADE}{semantic pyramid anomaly detection}{
	TODO
}

\newdualentry{pq}{PQ}{panoptic quality}{
	TODO
}

\newdualentry{sq}{SQ}{segmentation quality}{
	TODO
}

\newdualentry{rq}{RQ}{recognition quality}{
	TODO
}

\newdualentry{fcdd}{FCDD}{fully convolutional data descriptor}{
	TODO
}

\newdualentry{padim}{PaDiM}{patch distribution modeling}{
	TODO
}

\newglossaryentry{rgb}{name={RGB},description={
	Standard color model in images. RGB stands for \enquote{red, green, blue} which the three color values indicate
}}

\newglossaryentry{hsv}{name={HSV},description={
	Alternative color model in images, better resembling human perceiving of color. HSV stands for \enquote{hue, saturation, value}
}}

\newglossaryentry{adam}{name={Adam}, description={
	An alternative to \gls{sgd} with better properties
}}

\newglossaryentry{led}{name={LED}, description={
	LED stands for Light Emitting Diode
}}

\newglossaryentry{umap}{name={UMAP}, description={
	\enquote{Uniform Manifold Approximation and Projection (UMAP) is a dimension reduction technique that can be used for visualisation similarly to \gls{tsne}, but also for general non-linear dimension reduction.}\cite{McInnes2018UmapUniformmanifold}
}}
\makeglossaries
\glsdisablehyper

\usepackage{url}

\hyphenation{op-tical net-works semi-conduc-tor}
\usepackage{csquotes}

\usepackage{pgfkeys-peek}
\usepackage[export]{adjustbox}

\newcommand{\etal}{et al.\ }
\newcommand{\ie}{i.e.\ }
\newcommand{\eg}{e.g.\ }

\usepackage[final]{ifdraft} %
\usepackage{calc}
\usepackage{multicol}
\usepackage{pslatex}
\usepackage{apalike}
\usepackage{epsfig}
\newcommand{\boldentry}[2]{%
\multicolumn{1}{S[table-format=#1,
mode=text,
text-rm=\fontseries{b}\selectfont
]}{#2}}

\newcommand{\boldentrynospace}[2]{%
\multicolumn{1}{S[table-format=#1,
mode=text,
text-rm=\fontseries{b}\selectfont
]@{}}{#2}}
\usepackage{./template/scitepress/SCITEPRESS}

\begin{document}
\title{Transfer Learning Gaussian Anomaly Detection by Fine-tuning Representations}

\ifdraft{\author{Redacted for double-blind review}}{
	\author{\authorname{Oliver Rippel\sup{1}\orcidAuthor{0000-0002-4556-5094}, Arnav Chavan\sup{2}, Chucai Lei\sup{1}, and Dorit Merhof\sup{1}\orcidAuthor{0000-0002-1672-2185}} %
		\affiliation{\sup{1}Institute of Imaging \& Computer Vision, RWTH Aachen University, Aachen, Germany}
		\affiliation{\sup{2}Indian Institute of Technology, ISM Dhanbad, India}
		\email{oliver.rippel@lfb.rwth-aachen.de}
	}
}

\ifdraft{
\hypersetup{
	pdftitle    = {Transfer Learning Gaussian Anomaly Detection by Fine-tuning Representations},
	pdfauthor = {Redacted for double-blind review},
	pdfkeywords = {Anomaly Detection, Anomaly Segmentation, Transfer Learning, PDF Estimation, Visual Inspection}
}
}
{
\hypersetup{
pdftitle    = {Transfer Learning Gaussian Anomaly Detection by Fine-tuning Representations},
pdfauthor = {Oliver Rippel, Arnav Chavan, Chucai Lei, Dorit Merhof},
pdfkeywords = {Anomaly Detection, Anomaly Segmentation, Transfer Learning, PDF Estimation, Visual Inspection}
}
}

\abstract{
	Current state-of-the-art \gls{ad} methods exploit the powerful representations yielded by large-scale ImageNet training.
	However, catastrophic forgetting prevents the successful fine-tuning of pre-trained representations on new datasets in the semi-supervised setting, and representations are therefore commonly fixed.
	In our work, we propose a new method to overcome catastrophic forgetting and thus successfully fine-tune pre-trained representations for \gls{ad} in the transfer learning setting.
	Specifically, we induce a multivariate Gaussian distribution for the normal class based on the linkage between generative and discriminative modeling, and use the Mahalanobis distance of normal images to the estimated distribution as training objective.
	We additionally propose to use augmentations commonly employed for \acrlong{vrm} in a validation scheme to detect onset of catastrophic forgetting. %
	Extensive evaluations on the public MVTec dataset reveal that a new state of the art is achieved by our method in the \gls{ad} task while simultaneously achieving \acrlong{as} performance comparable to prior state of the art. %
	Further, ablation studies demonstrate the importance of the induced Gaussian distribution as well as the robustness of the proposed fine-tuning scheme with respect to the choice of augmentations. %
}

\keywords{Anomaly Detection, Anomaly Segmentation, Transfer Learning, PDF Estimation, Visual Inspection}

\onecolumn \maketitle \normalsize \setcounter{footnote}{0} \vfill

\glsresetall

\section{\uppercase{Introduction}}
\Gls{ad} in images is concerned with finding images that deviate from a prior-defined concept of normality, and poses a fundamental computer vision problem with application domains ranging from industrial quality control \cite{Bergmann2021MVTecAnomalyDetection} to medical image analysis~\cite{Schlegl2019fAnoGANFast}.
Extending \gls{ad}, \gls{as} tries to identify the visual patterns inside anomalous images that constitute the anomaly.
In general, \gls{ad}/\gls{as} tasks are defined by the following two characteristics\footnote{For a general, exhaustive overview of \gls{ad} we refer the reader to~\cite{Ruff2021UnifyingReviewDeep}.}:

\begin{enumerate}
	\item Anomalies are rare events, \ie their prevalence in the application domain is low.
	\item There exists limited knowledge about the anomaly distribution, \ie it is ill-defined.
\end{enumerate}
Together, these characteristics result in \gls{ad}/\gls{as} datasets that are heavily imbalanced, often containing only few anomalies for model verification and testing.

As a consequence, \gls{ad}/\gls{as} algorithms focus on the semi-supervised setting, where exclusively normal data is used to establish a model of normality~\cite{Bergmann2021MVTecAnomalyDetection,Schlegl2019fAnoGANFast,Ruff2018DeepOneClass}.
Since learning discriminative representations from scratch is difficult \cite{Rippel2021ModelingDistributionNormal}, state-of-the-art methods leverage representation gained by training on ImageNet \cite{Deng2009ImageNetlargescale} as the basis for \gls{ad}/\gls{as} \cite{Rippel2021GaussianAnomalyDetection,Defard2020PaDiMPatchDistribution,Andrews2016Transferrepresentationlearning,Bergmann2020UninformedstudentsStudent,Cohen2020SubImageAnomaly,Christiansen2016DeepAnomalyCombiningbackground,Perera2019Learningdeepfeatures,Li2021CutPasteSelfSupervised}. %

Fine-tuning these representations on the dataset at hand now offers the potential of additional performance improvements.
However, fine-tuning is hindered by catastrophic forgetting, which is defined as the loss of discriminative features initially present in the model in context of \gls{ad} \cite{Deecke2021TransferBasedSemantic}.
In fact, fine-tuning is so difficult that it is simply foregone in the majority of \gls{ad}/\gls{as} approaches (refer also \autoref{subsec:fixed_repr}).
While methods have been proposed to tackle catastrophic forgetting \cite{Reiss2021PANDAAdaptingPretrained,Deecke2020DeepAnomalyDetection,Perera2019Learningdeepfeatures,Liznerski2021ExplainableDeepOne,Ruff2020RethinkingAssumptionsDeep,Deecke2021TransferBasedSemantic}, they currently ignore the feature correlations inherent to the pre-trained networks (refer \autoref{fig:overview_method}, \cite{Rodriguez2017RegularizingCNNsLocally,Ayinde2019RegularizingDeepNeural}).
Moreover, many of the aforementioned methods employ synthetic anomalies as surrogates for the true anomaly distribution to generate a supervised loss, a concept called \gls{oe} (refer \autoref{fig:overview_method}) \cite{Hendrycks2019DeepAnomalyDetection}.
While performance gains have been reported here, a significant dataset bias is also induced by \gls{oe} \cite{Ye2021UnderstandingEffectBias}. 

Our contributions are as follows:
\begin{itemize}
	\item Based on findings from \cite{Lee2018simpleunifiedframework} and \cite{Rippel2021ModelingDistributionNormal}, we induce a multivariate Gaussian for the normal data distribution in the transfer learning setting, incorporating the feature correlations inherent to the pre-trained network.
	      We thus propose to fine-tune the pre-trained model by minimizing the Mahalanobis distance \cite{Mahalanobis1936generalizeddistancestatistics} to the estimated Gaussian distribution using normal data only, and forego \gls{oe} to avoid incorporating an additional dataset bias.
		  We furthermore show how the proposed objective relates to and extends the prior-used optimization functions.
	\item We propose an early stopping criterion based on data augmentations commonly used in semi-supervised learning for \gls{vrm} \cite{Chapelle2001VicinalRiskMinimization,Hendrycks2020AugMixSimpleData} to detect onset of catastrophic forgetting. 
	\item We demonstrate the effectiveness of both proposals using the public MVTec dataset \cite{Bergmann2021MVTecAnomalyDetection}, and perform extensive ablation studies to investigate the sensitivity of our approach with respect to the chosen augmentation schemes.
\end{itemize}

\begin{figure}[tbp]
	\centering
	\includegraphics{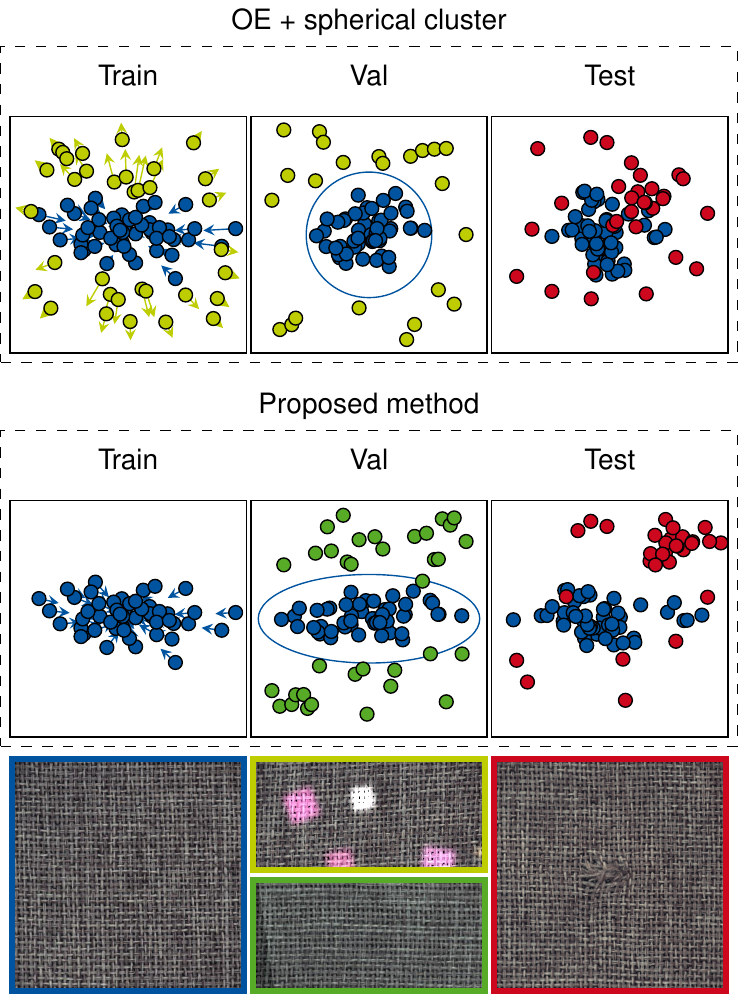}
	\caption[Comparison of the proposed method and related fine-tuning approaches.]{Comparison of the proposed method and related fine-tuning approaches.
		Anomalies (shown in red) are often subtle, and state-of-the-art methods use either them or synthetic anomalies (shown in yellow) for \gls{oe}.
		However, doing so introduces a significant bias \cite{Ye2021UnderstandingEffectBias}.
		Moreover, state-of-the-art fine-tuning methods ignore the feature correlations inherent to the pre-trained network \cite{Rodriguez2017RegularizingCNNsLocally,Ayinde2019RegularizingDeepNeural}.
		In the proposed method, these correlations are taken into account by means of the Gaussian assumption.
		Furthermore, we argue that augmentations used for \gls{vrm} (shown in green) are well-suited for detecting the onset of catastrophic forgetting, and can thus be used to validate semi-supervised training that uses normal data only (shown in blue).
		Thereby, the bias induced by \gls{oe} is reduced/circumvented.
		A dot in the scatter plot corresponds to a complete image.}
	\label{fig:overview_method}
\end{figure}

\section{\uppercase{Related Work}}
Features generated by large-scale dataset training (\eg ImageNet) are commonly employed in literature to achieve \gls{ad}/\gls{as} on new tasks in a transfer learning setting.

\subsection{Fixed representations}\label{subsec:fixed_repr}
The representations of the pre-trained network are commonly fixed to prevent catastrophic forgetting.
To nevertheless improve \gls{ad}/\gls{as} performance even with fixed representations, Bergmann \etal \cite{Bergmann2020UninformedstudentsStudent} employ a two-stage knowledge distillation framework to achieve \gls{ad} and \gls{as} in a transfer learning setting.
Here, they directly regress the intermediate representations of a ResNet18~\cite{He2016Deepresiduallearning} pre-trained on ImageNet.
Furthermore, Rudolph \etal \cite{Rudolph2021SameSameDifferNet} fit an unconstrained probability distribution by means of Normalizing Flows \cite{Rezende2015VariationalInferenceNormalizing} to the nominal class. %

Moreover, features from pre-trained networks are also used as the basis for classical \gls{ad} methods.
For example, Andrews \etal \cite{Andrews2016Transferrepresentationlearning} fit a discriminative \acrlong{ocsvm} \cite{Schoelkopf2001Estimatingsupporthigh} to features extracted from intermediate layers of a VGG \cite{Simonyan2015VeryDeepConvolutional} network. %
Furthermore, $k$-NN has also been used to realize \gls{ad}/\gls{as} on pre-trained features \cite{Cohen2020SubImageAnomaly}.
Last, generative algorithms such as Gaussian \gls{ad} \cite{Christiansen2016DeepAnomalyCombiningbackground,Sabokrou2018DeepanomalyFully,Defard2020PaDiMPatchDistribution,Rippel2021ModelingDistributionNormal} are also commonly employed.

\subsection{Fine-tuning representations}
Even though it offers potential benefits, fine-tuning approaches are limited by onset of catastrophic forgetting.
In contrast to the typical continual learning scheme, catastrophic forgetting in \gls{ad}/\gls{as} is incurred by the unavailability of anomalous data.
As a consequence, discriminative feature combinations initially present in the pre-trained network are lost during fine-tuning, since they are most likely absent/missing in the normal data \cite{Tax2003Featureextractionone,Rippel2021ModelingDistributionNormal}.
Ultimately this leads to a reduced \gls{ad}/\gls{as} performance.

Due to the absence of anomalies, fine-tuning approaches base their learning objective on the \textit{concentration} assumption instead, \ie they try to find a compact description of the normal class in high-dimensional representations.
The most commonly used optimization formulation for this is the Deep \gls{svdd} objective \cite{Ruff2018DeepOneClass}, defined as
\begin{equation}
	\label{eq:svdd}
	\min_{\mathcal{W}} \frac{1}{n} \sum_{i=1}^{n}\norm{\Phi(\vec{x}_i; \mathcal{W}) - \vec{c}}_1 + \frac{\lambda}{2}\sum_{l=1}^{L}\norm{\vec{W}^l}^2_F.
\end{equation}
Here, $\Phi: \mathbb{R}^{C \times H \times W} \to \mathbb{R}^{D}$ is a neural network parametrized by its weights $\mathcal{W} = \{\vec{W}^1, ..., \vec{W}^L\}$, and a hypersphere is minimized around the cluster center $\vec{c}$ subject to $L_2$ weight regularization, with $\norm{\cdot}_F$ denoting the Frobenius norm.

To now overcome catastrophic forgetting, three main procedures have been proposed in literature:
(I) Perera \& Patel \cite{Perera2019Learningdeepfeatures} jointly optimize \autoref{eq:svdd} together with the original task to ensure that the original feature discriminativeness does not deteriorate.
Specifically, they use the arithmetic mean as $\vec{c}$ and $L_2$ norm over the $L_1$ norm in \autoref{eq:svdd}, and perform joint optimization with ImageNet-1k training.
Joint optimization, however, requires access to the original dataset, which may not always be feasible.
(II) Reiss \etal \cite{Reiss2021PANDAAdaptingPretrained} propose to make use of \gls{ewc}\cite{Kirkpatrick2017Overcomingcatastrophicforgetting}, a technique proposed in continual learning, to overcome catastrophic forgetting.
This method, however, also requires access to the original dataset, which may again not always be feasible.
(III) Deecke \etal \cite{Deecke2021TransferBasedSemantic} propose to make use of $L_2$ regularization with respect to the initial, pre-trained weights, arguing that only subtle feature adaptations should be necessary when fine-tuning.
Alternatively, they also propose to learn only a modulation of frozen features by means of newly introduced, residual adaptation layers, instead of tuning them directly.

All the aforementioned techniques can be applied in combination with \gls{oe} \cite{Hendrycks2019DeepAnomalyDetection}.
Here, either synthetic anomalies or datasets disjoint to the target domain are used as surrogate anomalies to facilitate training of supervised, discriminative models.
Ruff \etal \cite{Ruff2020RethinkingAssumptionsDeep} formulate a \gls{hsc} to incorporate \gls{oe} to the Deep \gls{svdd} objective.
Specifically, they optimize
\begin{equation}
	\label{eq:hsc}
	\begin{split}
		\min_{\mathcal{W}} & \frac{1}{n} \sum_{i=1}^{n}y_i\norm{\Phi(\vec{x}_i; \mathcal{W})}^2 \\
		& - (1 - y_i) \log\left(1 - \exp(-\norm{\Phi(\vec{x}_i; \mathcal{W})}^2\right),
	\end{split}
\end{equation}
where $y_i$ denotes whether a sample is either normal ($y_i = 1$) or anomalous ($y_i = 0$).
While generalization to anomalies unseen during training has been demonstrated for \gls{oe} \cite{Hendrycks2019DeepAnomalyDetection}, a bias is undoubtedly introduced by this method.
Specifically, Ye \etal \cite{Ye2021UnderstandingEffectBias} show that labeled anomalies differing subtly from the normal class have a large impact in \gls{oe}, possibly degrading performance on unseen anomaly types that are also similar to the normal class.
They further show that empirical gains can only be guaranteed when anomalies are used for \gls{oe} that differ strongly from the normal class instead.

Nevertheless, gains have been reported by approaches that incorporate \gls{oe}, such as CutPaste \cite{Li2021CutPasteSelfSupervised} or \gls{fcdd} \cite{Liznerski2021ExplainableDeepOne}.
Here, CutPaste follows a two-stage procedure, \ie fine-tuning a supervised classifier by means of \gls{oe} with carefully crafted, synthetic anomalies followed by Gaussian \gls{ad} as proposed in \cite{Rippel2021ModelingDistributionNormal}.
\Gls{fcdd} applies the \gls{hsc} objective to spatial feature maps, \ie intermediate representations of a pre-trained network that still possess spatial dimensions, to fine-tune the network directly. %
Similar to CutPaste, they also use carefully crafted, synthetic anomalies for \gls{oe}.

Out of all related works, only \gls{fcdd} and CutPaste evaluate performance on a dataset where normal and anomalous classes differ subtly, namely the MVTec dataset \cite{Bergmann2021MVTecAnomalyDetection}. %

\section{\uppercase{Transfer Learning Gaussian Anomaly Detection}}
In our work, we propose to fine-tune the representations of pre-trained networks based on the Gaussian assumption.
A motivation behind the seemingly overly simplistic Gaussian assumption can be inferred from Lee \etal \cite{Lee2018simpleunifiedframework}.
Here, authors have induced a \gls{gda} for \gls{ood} detection.
Investigations into the unreasonable effectiveness of the Gaussian assumption by Kamoi and Kobayashi \cite{Kamoi2020WhyisMahalanobis} have revealed that feature combinations containing low variance for the normal/nominal class are ultimately those discriminative to the \gls{ood} data.
Independent investigations into the same phenomenon for \gls{ad} by Rippel \etal \cite{Rippel2021ModelingDistributionNormal} have revealed the same finding for the transfer learning setting.
However, despite its elegant simplicity and outstanding performance, the Gaussian assumption has not yet been used to fine-tune pre-trained features for \gls{ad}.

The Gaussian distribution is given by %
\begin{equation}
	\varphi_{\vec{\mu},\Sigma}(\vec{x}) := \frac{1}{\sqrt{(2\pi)^D} \abs{\det\Sigma}} e^{-\frac{1}{2} (\vec{x} - \vec{\mu})^\top \Sigma^{-1} (\vec{x} - \vec{\mu})},
\end{equation}
with $D$ being the number of dimensions, $\vec{\mu} \in \mathbb{R}^D$ being the mean vector and $\Sigma \in \mathbb{R}^{D\times D}$ being the symmetric covariance matrix of the distribution, which must be positive definite.

Under a Gaussian distribution, a distance measure between a particular point $\vec{x} \in \mathbb{R}^D$ and the distribution is called the Mahalanobis distance \cite{Mahalanobis1936generalizeddistancestatistics}, given as %
\begin{equation} \label{eq:mahalanobis}
	M(\vec{x}) = \sqrt{\left(\vec{x} - \vec{\mu}\right)^\top \Sigma^{-1} \left(\vec{x} - \vec{\mu}\right)}.
\end{equation}
Let us compare the Mahalanobis distance (\autoref{eq:mahalanobis}) to the Deep \gls{svdd} objective (\autoref{eq:svdd}).
When imposing a univariate Gaussian with zero-mean and unit-variance per feature, the Mahalanobis distance reduces to the $L_2$ distance to the mean, recapitulating the original Deep \gls{svdd} objective with $L_2$ norm instead of $L_1$ norm.
In other words, minimizing the  Deep \gls{svdd} objective in conjunction with $L_2$ norm \textit{implicitly assumes} that features follow independent, univariate Gaussians with zero-mean and unit-variance.
This assumption, however, is in disagreement with findings in literature, where strong correlations have been observed across deep features extracted from \glspl{cnn} \cite{Rodriguez2017RegularizingCNNsLocally,Ayinde2019RegularizingDeepNeural}.
Our induced multivariate Gaussian prior thus actually imposes a lesser inductive bias than the Deep \gls{svdd} objective, and allows for more flexible distributions.
Internal experiments furthermore showed that identical performance could be achieved when decorrelating the pre-trained features by means of Cholesky decomposition prior to training them with the Deep \gls{svdd} objective.

Since others \cite{Cohen2020SubImageAnomaly,Defard2020PaDiMPatchDistribution,Liznerski2021ExplainableDeepOne} have demonstrated that good \gls{as} performance can be achieved by utilizing intermediate feature representations that maintain spatial dimensions, we propose to apply the Mahalanobis distance to Gaussians fitted to intermediate feature representations as well.
Specifically, let $\Phi: \mathbb{R}^{C \times H \times W} \to \mathbb{R}^{D}$ be a \gls{cnn} with its intermediate mappings denoted as $\Phi_m : \mathbb{R}^{C_{m-1} \times H_{m-1} \times W_{m-1}} \to \mathbb{R}^{C_{m} \times H_{m} \times W_{m}}$.
Then, the Mahalanobis distance of $\Phi_m$ is a  given by applying \autoref{eq:mahalanobis} to each spatial element independently, yielding a matrix $\vec{A}_m$ of size $H_m \times W_m$. %
Note that we account for small local changes in image composition by modeling $\vec{\mu}$ independently per location \& tie $\Sigma$ across locations.
This has already been shown to be particularly powerful for fixed features in \cite{Rippel2021AnomalyDetectionAutomated}, and we perform an evaluation of its effectiveness in \autoref{subsubsec:ablations}. %
Furthermore, we use $max(\vec{A})$ to give an anomaly score per level $m$ of $\Phi$, which is also used for optimization.
Using $max(\vec{A})$ over $mean(\vec{A})$ is motivated by \cite{Defard2020PaDiMPatchDistribution}, which have shown that strong \gls{ad} results can be achieved by aggregating in such a manner for fixed representations.
We argue that, when fine-tuning \gls{ad}, one wants to modify exactly those feature combinations that are currently most perceived to be anomalous for normal images, in order to ensure that the true data distribution fits the estimated Gaussian distribution better, and assess its effects in \autoref{subsubsec:ablations}. %

Our overall minimization objective is thus
\begin{equation}
	\label{eq:loss_objective}
	\min_{\mathcal{W}} \frac{1}{n \cdot m} \sum_{i=1}^{n} \sum_{}^{m} max(\vec{A}_m(\Phi(\vec{x}_i; \mathcal{W}))). %
\end{equation}
Note that the parameters $\vec{\mu}_m$ and $\Sigma_m$ of the Gaussians are not learned.
Instead, we use the empirical mean $\vec{\mu}_m$ and estimate $\Sigma_m$ using shrinkage as proposed by Ledoit \etal \cite{Ledoit2004wellconditionedestimator}, and leave them fixed afterwards.
This yields a stronger bias and was shown to prevent catastrophic forgetting in prior experiments. %

For \gls{as}, we propose to upsample $\vec{A}_m$ using bilinear interpolation similarly to \cite{Cohen2020SubImageAnomaly,Defard2020PaDiMPatchDistribution} over Gaussian interpolation.
Bilinear interpolation is free of hyperparameters and thus more robust while simultaneously offering competitive results (selection of the Gaussian interpolation kernel was shown to have a strong effect on \gls{as} performance in \cite{Liznerski2021ExplainableDeepOne}). %
After their respective upsampling, pixel-wise averaging across all heatmaps yields the overall anomaly score map.

Note that while \gls{oe} could be easily integrated into our proposed approach by maximizing \autoref{eq:loss_objective} for synthetic anomalies, we forego it to avoid the induction of an additional bias incurred by providing surrogates for the anomaly distribution \cite{Ye2021UnderstandingEffectBias}.

\subsection{Early stopping via \gls{vrm}}\label{subsec:early_stopping_vrm}
While we forego \gls{oe}, we still need to detect onset of catastrophic forgetting.
To this end, we propose to approximate the compactness/discriminativeness of the learned distribution.
Specifically, we argue that a good model should be able to distinguish between subtle variations of the normal data and the normal data itself as a surrogate for \gls{ad} performance.
We therefore sample the vicinity of the normal data distribution by employing augmentations used for \gls{vrm} in semi-supervised learning schemes (refer \autoref{fig:overview_method}).
Afterwards, we measure the capability of a tuned network $\Phi$ to distinguish between subtle vicinal variations and normal data by means of \gls{auroc} as surrogate of the model's discriminativeness. %
We select both best model state based on this criterion and perform early stopping should \gls{auroc} no longer improve.
As multiple augmentation schemes  have been proposed for \gls{vrm} in literature, we compare \textit{AugMix} \cite{Hendrycks2020AugMixSimpleData} with \textit{CutOut} \cite{DeVries2017ImprovedRegularizationConvolutional} to investigate the dependence of early stopping on augmentation types. %
Additionally, we compare with \textit{Confetti} noise, but note that this has been introduced as a surrogate for the anomaly distribution originally in \cite{Liznerski2021ExplainableDeepOne}.

We furthermore note that validating the model by assessing its capability to distinguish between subtle variations and normal data could introduce an additional bias.
We argue however, that this bias should be less strong than when using these subtle variations for \gls{oe} directly, and show this to be the case in \autoref{subsec:ad}.

\section{\uppercase{Experiments}}
First, we briefly introduce the dataset used for evaluating our approach as well as the employed metrics.

\subsection{Dataset and Evaluation metrics}
We use the public MVTec dataset \cite{Bergmann2021MVTecAnomalyDetection} to test and compare our approach with literature.
The reasons for this are two-fold:
First, MVTec consists of subtle anomalies in high-resolution images as opposed to the \gls{ad} tasks commonly constructed from classification datasets (\eg one-versus-rest based on CIFAR-10 \cite{Liznerski2021ExplainableDeepOne}).
Therefore, MVTec is more challenging and indicative of real-world \gls{ad}/\gls{as} performance.
Second, MVTec provides binary segmentation masks that can be used to evaluate \gls{as} performance, which is infeasible for classification datasets.
MVTec itself consists of 15 industrial product categories in total (10 object and 5 texture classes), and contains 5354 images overall.

To evaluate \gls{ad} performance, we report the \gls{auroc}, a metric commonly used to evaluate binary classifiers \cite{Ferri2011coherentinterpretationAUC}.
For \gls{as}, we report the pixel-wise \gls{auroc} as well as the \gls{pro} curve until 30\% \gls{fpr} as proposed in \cite{Bergmann2020UninformedstudentsStudent}.
While pixel-wise \gls{auroc} represents an algorithm's capability of identifying anomalous pixels, \gls{pro} focuses more on an algorithm's performance at detecting small, locally constrained anomalies.

\subsection{Anomaly Detection}\label{subsec:ad}
We first assess whether our proposed fine-tuning approach improves \gls{ad} performance.

\paragraph{Training \& Evaluation details.}
We perform a 5-fold evaluation over the training set of each MVTec category, training an EfficientNet-B4 \cite{Tan2019EfficientNetRethinkingModel} to minimize the objective given by \autoref{eq:loss_objective}, and split the training dataset into 80\% used for training and 20\% used for validation.
We choose EfficientNet based on its strong ImageNet performance \cite{Tan2019EfficientNetRethinkingModel}, as architectures with stronger ImageNet performance have been shown to yield better features for transfer learning in \cite{Kornblith2019Dobetterimagenet}, which has been further confirmed for transfer learning in \gls{ad} \cite{Rippel2021ModelingDistributionNormal}.
Moreover, EfficientNet-B4s have been shown to offer a good trade-off between model complexity and \gls{ad} performance \cite{Rippel2021ModelingDistributionNormal}.
To demonstrate the general applicability of our approach, we omit the selection of best-performing feature levels, simply extracting the features from every \enquote{level} of the EfficientNet as denoted in \cite{Tan2019EfficientNetRethinkingModel}. %
We also train models to minimize the Deep \gls{svdd} (Equation \ref{eq:svdd}) and \gls{fcdd} objectives.
Here, we apply our early stopping criterion to both objectives, and simultaneously use the synthetic anomalies for \gls{oe} in the \gls{fcdd} objective. %
This is done to factor out differences in model performance yielded by changing the underlying feature extractor (to the best of our knowledge, Deep \gls{svdd} and \gls{fcdd} objectives have not yet been applied to EfficientNets for transfer learning \gls{ad} in the fine-tuning setting).
All models are trained using a batch size of 8, \gls{adam} optimizer \cite{Kingma2015AdamMethodStochastic} and a learning rate of $\num{1e-6}$.
The small learning rate is motivated by the fact that the feature representations are already discriminative as is (refer \eg to the fixed baseline in \autoref{tab:AD_results}), and only need to be tuned slightly.
Furthermore, BatchNormalization \cite{Ioffe2015BatchNormalizationAccelerating} statistics were kept frozen, as dataset sizes are too small to re-estimate them reliably.
Training was stopped when validation \gls{auroc} performance did not improve for 20 epochs, and 70\% of validation data was augmented and marked as anomalous.
We randomly sample 10 times the length of our validation dataset for validation to better approximate overall population characteristics, and sample augmentations from all \gls{vrm}-schemes with equal probability (refer also \autoref{subsubsec:ablations}).
In all experiments, we report $\mu \pm$ \gls{serr} aggregated over the 15 MVTec categories.

\begin{table*}[tbp]
	\caption{Comparison to the state of the art for \gls{ad}.
		We report $\mu \pm \text{\gls{serr}}$ \gls{auroc} scores in percent aggregated over all MVTec categories.
		Note that values reported for GeoTrans and GANomaly were taken from \cite{Fei2020Attributerestorationframework}, and values reported for US from \cite{Zavrtanik2020Reconstructioninpaintingvisual}.
		Abbreviations: \gls{serr} = \acrlong{serr}.
	}
	\label{tab:AD_results}
	\centering
	\glsunset{spade}
	\glsunset{padim}
	\begin{tabular}{@{}p{0em}@{}llS[table-format=2.1]S[table-format=1.1]@{}}
		\toprule
		 & Approach                                                      & Feature type & {Mean $\uparrow$} & {\gls{serr} $\downarrow$} \\
		\midrule
		 & GeoTrans \cite{Golan2018Deepanomalydetection}                 & Scratch      & 67.2              & 4.7                      \\
		 & GANomaly \cite{Akcay2018GANomalySemisupervised}               & Scratch      & 76.1              & 1.6                      \\
		 & ARNet \cite{Fei2020Attributerestorationframework}             & Scratch      & 83.9              & 2.8                      \\
		 & RIAD \cite{Zavrtanik2020Reconstructioninpaintingvisual}       & Scratch      & 91.7              & 1.8                      \\
		 & US \cite{Bergmann2020UninformedstudentsStudent}               & Scratch      & 87.7              & 2.8                      \\
		 & \gls{spade} \cite{Cohen2020SubImageAnomaly}                   & Frozen       & 85.5              & {---}                    \\
		 & Differnet \cite{Rudolph2021SameSameDifferNet}                 & Frozen       & 94.7              & 1.3                      \\
		 & \gls{padim} \cite{Defard2020PaDiMPatchDistribution}           & Frozen       & 95.3              & {---}                    \\ %
		 & Patch \gls{svdd} \cite{Yi2020PatchSVDDPatch}                  & Scratch      & 92.1              & 1.7                      \\
		 & Triplet Networks \cite{Tayeh2020DistanceBasedAnomaly}         & Scratch      & 94.9              & 1.2                      \\
		 & CutPaste \cite{Li2021CutPasteSelfSupervised}                  & Tuned        & 96.6              & 1.1                      \\
		 & Gaussian \gls{ad} \cite{Rippel2021ModelingDistributionNormal} & Frozen       & 95.8              & 1.2                      \\
		 \midrule
		 & Gaussian fine-tune (ours)                                     & Tuned        & \boldentry{2.1}{97.1} & \boldentrynospace{1.1}{0.8}            \\ %
		 &                                                               & Frozen       & 96.3              & 1.1                      \\
		 & Deep \gls{svdd}                                               & Tuned        & 90.5              & 2.7                      \\ %
		 &                                                               & Frozen       & 86.7              & 3.5                      \\ %
		 & \gls{fcdd}                                                    & Tuned        & 87.0              & 3.6                      \\ %
		\bottomrule
	\end{tabular}
\end{table*}

\paragraph{Results.} Analyzing results (\autoref{tab:AD_results}), it can be seen that our proposed transfer learning scheme improves over the baseline (frozen features), increasing \gls{ad} performance from $96.3 \pm 1.1$ to $97.1 \pm 0.8$ \gls{auroc} and setting a new state-of-the-art in \gls{ad} on the public MVTec dataset.
Furthermore, it can be seen that the Gaussian assumption is important, since \gls{svdd} and \gls{fcdd} objectives are already outperformed by our baseline where no fine-tuning of feature representations occurs ($90.5 \pm 2.7$  and $87.0 \pm 3.6$ vs.\ $96.3 \pm 1.1$). %
Still, it should be noted that fine-tuning using our proposed early stopping criterion improves results also here ($86.7 \pm 3.5$ to $90.5 \pm 2.7$ and $87.0 \pm 3.6$). %
Comparing Deep \gls{svdd} and \gls{fcdd} performances, it can be seen that worse \gls{ad} performance is achieved when using the subtle variations for \gls{oe} rather than just using them for model validation.

\subsubsection{Ablation studies}\label{subsubsec:ablations}
Next, we perform ablation studies to quantify the effects of proposed early stopping procedure, parametrization of the Gaussian distribution + aggregation of spatial anomaly scores on \gls{ad} performance.

\paragraph{Proposed early stopping procedure.} We perform two different experiments to assess effects of the proposed early stopping procedure.

First, we investigate effects of early stopping itself, comparing the proposed approach to both (I) the training for a fixed number of epochs (250 specifically) and (II) the early stopping without sampling the vicinity of the normal data distribution. %
Here, we use the minimum loss of \autoref{eq:loss_objective} over the validation set as an early stopping criterion and omit sampling the vicinity, focussing only on how well the normal data fits the distribution. %
\begin{table}[tbp]
	\caption{Effects of early stopping criterion on fine-tuning \gls{ad} performance.}
	\label{tab:early_stopping_AD}
	\centering
	\begin{tabular}{@{}lS[table-format=2.1]S[table-format=1.1]@{}}
		\toprule
		Method                          & {Mean $\uparrow$} & {\gls{serr} $\downarrow$} \\
		\midrule
		Vicinity sampling via \gls{vrm} & 97.1              & 0.9                      \\
		Fixed epochs                    & 96.3              & 1.2                      \\ %
		Validation loss                 & 95.4              & 1.2                      \\ %
		\midrule
		No Training                     & 96.3              & 1.1                      \\ %
		\bottomrule
	\end{tabular}
\end{table}

Assessing results (\autoref{tab:early_stopping_AD}), it can be seen that our proposed early stopping criterion is the only method that improves results.
Furthermore, early stopping based on validation loss alone leads to worse results than training for a fixed number of epochs.
Thus, validation loss alone is unable to detect onset of catastrophic forgetting.
It should also be noted that manually tuning the number of fixed epochs for training could have eventually led to improved results. %
Such a tuning, however, would need to be performed in a dataset specific manner and is not needed by our proposed early stopping criterion.

Second, we assess the method's sensitivity with respect to the chosen \gls{vrm}-type.
To this end, we perform an additional ablation study, varying the severity of \textit{AugMix} \cite{Hendrycks2020AugMixSimpleData} and investigate the suitability of \textit{CutOut} \cite{DeVries2017ImprovedRegularizationConvolutional} and \textit{Confetti noise} \cite{Liznerski2021ExplainableDeepOne}. %
We sample the depth of \textit{AugMix} uniformly from [1, 3], as further increasing the depth produced too strong variations in image appearance.
We also investigate benefits yielded by combining different augmentation schemes, where the augmentation of a sample is drawn randomly from $\{\textit{AugMix}, \textit{CutOut}, \textit{Confetti}\}$ with equal probability.
A reference image for each synthesis procedure is shown in \autoref{fig:synthetic_anomalies_ref}.

\begin{figure}
	\centering
	\subfloat[]{\includegraphics[width=0.32\linewidth]{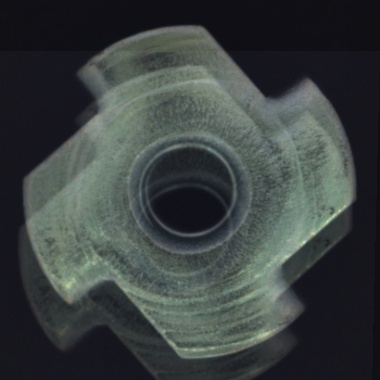}}
	\vspace{\fill}
	\subfloat[]{\includegraphics[width=0.32\linewidth]{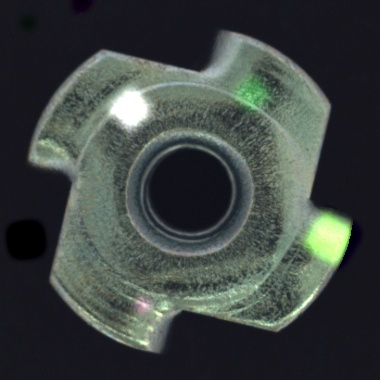}}
	\vspace{\fill}
	\subfloat[]{\includegraphics[width=0.32\linewidth]{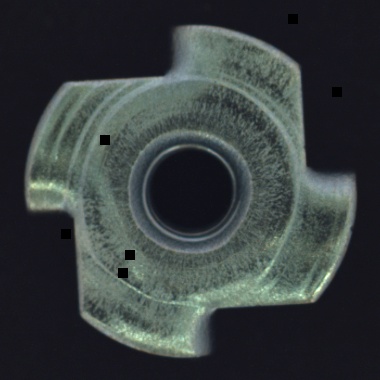}}
	\caption{Reference synthesis results generated by (a) \textit{AugMix}, (b) \textit{Confetti} and (c) \textit{CutOut} for the metal nut class.}
	\label{fig:synthetic_anomalies_ref}
\end{figure}

\begin{table}[tbp]
	\caption{Effect of \gls{vrm}-type on \gls{ad} performance.}
	\label{tab:ablation_vicinity}
	\centering
	\begin{tabular}{@{}llS[table-format=2.1]S[table-format=1.1]@{}}
		\toprule
		\multicolumn{2}{@{}l}{Method}                  & {Mean $\uparrow$} & {\gls{serr} $\downarrow$}       \\
		\midrule
		\multicolumn{2}{@{}l}{\textit{AugMix}}         &                   &                                \\ %
		                                               & sev = 3           & 97.1                     & 0.8 \\
		                                               & sev = 4           & 97.0                     & 0.9 \\
		                                               & sev = 5           & 97.0                     & 0.8 \\
		                                               & sev = 6           & 97.0                     & 0.8 \\ %
		                                               & sev = 7           & 97.1                     & 0.8 \\
		\multicolumn{2}{@{}l}{\textit{Confetti noise}} & 96.7              & 1.0                            \\ %
		\multicolumn{2}{@{}l}{\textit{CutOut}}         & 96.6              & 1.0                            \\ %
		\multicolumn{2}{@{}l}{\textit{All}}            & 97.1              & 0.8                            \\ %
		\bottomrule
	\end{tabular}
\end{table}

Assessing results (\autoref{tab:ablation_vicinity}), it can be seen that all augmentation schemes improve results over the baseline, where feature adaptation is omitted (\autoref{tab:AD_results}).
Furthermore, \textit{AugMix} outperforms the other two methods slightly, and the performance is invariant to the severity of applied augmentations.
Last, when jointly applying all augmentations, the same performance is reached as when applying only the best augmentation.
This indicates that augmentation schemes do not affect each other negatively, and reduces the complexity of the approach:
One can simply pool multiple augmentation strategies and still achieve the overall best performance.

\paragraph{Gaussian types \& aggregation.}
We also investigate the effects of both different distribution types fit to the normal class and spatial aggregation of anomaly scores on \gls{ad} performance.
Specifically, we compare a global Gaussian distribution (shared $\vec{\mu}$ and $\Sigma$ across all spatial locations of a \enquote{level}), a tied Gaussian distribution (individual $\vec{\mu}$ and tied $\Sigma$) as well as local Gaussian distributions (individual $\vec{\mu}$ and $\Sigma$ per location).
Except for the parametrization of the Gaussian, training details are identical as before. %
For the aggregation methods, we compare mean and maximum aggregation for generation of image-level \gls{ad} scores. %

\begin{table}[tbp]
	\caption{Effect of different types of distribution/aggregation methods on \gls{ad} performance.}
	\label{tab:ablation_gaussian_type}
	\centering
	\begin{tabular}{@{}llS[table-format=2.1]S[table-format=1.1]@{}}
		\toprule
		\multicolumn{2}{@{}l}{Method}                    & {Mean $\uparrow$}  & {\gls{serr} $\downarrow$}       \\
		\midrule
		\multicolumn{2}{@{}l}{Global Gaussian}           &                    &                                \\
		                                                 & aggregation = mean & 93.7                     & 2.3 \\ %
		                                                 & aggregation = max  & 96.9                     & 0.9 \\ %
		\multicolumn{2}{@{}l}{One Gaussian per location} &                    &                                \\
		                                                 & aggregation = mean & 97.3                     & 0.8 \\ %
		                                                 & aggregation = max  & 97.3                     & 0.8 \\ %
		\multicolumn{2}{@{}l}{Tied Gaussian}             &                    &                                \\
		                                                 & aggregation = mean & 93.0                     & 2.3 \\ %
		                                                 & aggregation = max  & 97.1                     & 0.8 \\ %
		\bottomrule
	\end{tabular}
\end{table}

When assessing results (\autoref{tab:ablation_gaussian_type}), it can be seen that maximum aggregation performs best across all methods.
Furthermore, effects of max aggregation are stronger for the global as well as the tied Gaussian distribution than the local Gaussian fit per location.
Overall, \gls{ad} performance of the local Gaussian model is best.
However, it also has the highest memory requirement, increasing the overall memory footprint of the method by a factor of 10. %
Notably, it also has the highest base \gls{ad} performance and only negligible gains are achieved by fine-tuning.
It is therefore infeasible in practice, and not regarded further.

\subsection{Anomaly Segmentation}
Since our approach maintains spatial resolution of anomaly scores, \gls{as} can also be achieved similarly to \cite{Liznerski2021ExplainableDeepOne,Defard2020PaDiMPatchDistribution}.
\paragraph{Training \& Evaluation details.}
To factor out effects of pre-trained model selection on \gls{as} performance and give a fair comparison to competing fine-tuning methods, we also report \gls{as} performance of EfficientNet-B4s fine-tuned with either \gls{fcdd} + \gls{oe} or Deep \gls{svdd}.
All other hyperparameters are same as in \autoref{subsec:ad}.

\begin{table*}[t]
	\caption{\gls{auroc} scores in percent for pixel-wise segmentation.
		We report $\mu \pm \text{\gls{serr}}$ across MVTec categories.
		Values reported for US were sourced	from \cite{Zavrtanik2020Reconstructioninpaintingvisual}.
	}
	\label{tab:AS_ROC}
	\centering
	\glsunset{vae}
	\begin{tabular}{@{}p{0em}@{}llS[table-format=2.1]S[table-format=1.1]@{}}
		\toprule
		 & Approach                                                                 & Feature type & {Mean $\uparrow$} & {\gls{serr} $\downarrow$} \\
		\midrule
		 & CAVGA-R\textsubscript{u} \cite{Venkataramanan2020AttentionGuidedAnomaly} & Scratch      & 89.0              & {---}                    \\ %
		 & RIAD \cite{Zavrtanik2020Reconstructioninpaintingvisual}                  & Scratch      & 94.2              & 1.3                      \\
		 & \gls{vae}-Attention \cite{Liu2020TowardsVisuallyExplaining}              & Scratch      & 86.1              & 2.3                      \\
		 & US \cite{Bergmann2020UninformedstudentsStudent}                          & Scratch      & 93.9              & 1.6                      \\
		 & \gls{spade} \cite{Cohen2020SubImageAnomaly}                              & Frozen       & 96.0              & 0.9                      \\ %
		 & \gls{padim} \cite{Defard2020PaDiMPatchDistribution}                      & Frozen       & \boldentry{2.1}{97.5}    & \boldentrynospace{1.1}{0.4}            \\
		 & Patch \gls{svdd} \cite{Yi2020PatchSVDDPatch}                             & Scratch      & 95.7              & 0.6                      \\ %
		 & \gls{fcdd} \cite{Liznerski2021ExplainableDeepOne}                        & Tuned        & 92.0              & {---}                    \\ %
		 & CutPaste \cite{Li2021CutPasteSelfSupervised}                             & Tuned        & 96.0              & 0.8                      \\
		 \midrule
		 & Gaussian fine-tune (ours)                                                & Tuned        & 96.5              & 0.8                      \\ %
		 &                                                                          & Frozen       & 96.4              & 0.7                      \\ %
		 & Deep \gls{svdd}                                                          & Tuned        & 95.1              & 0.9                      \\ %
		 &                                                                          & Frozen       & 95.2              & 0.9                      \\ %
		 & \gls{fcdd}                                                               & Tuned        & 96.2              & 0.6                      \\ %
		\bottomrule
	\end{tabular}
\end{table*}

\begin{table*}[t]
	\caption{
		Area under the \gls{pro} curve as defined by \cite{Bergmann2021MVTecAnomalyDetection,Bergmann2020UninformedstudentsStudent} up to a \acrlong{fpr} of 30\%.
		Values for state-of-the-art methods are directly taken from the corresponding sources.
		We report $\mu\pm$\gls{serr} values.
	}
	\label{tab:AS_pro}
	\centering
	\begin{tabular}{@{}p{0em}@{}llS[table-format=2.1]S[table-format=1.1]@{}}
		\toprule
		 & Approach                                            & Feature type & {Mean $\uparrow$} & {\gls{serr} $\downarrow$} \\
		\midrule
		 & \gls{spade} \cite{Cohen2020SubImageAnomaly}         & Frozen       & 91.7              & 1.4                      \\
		 & \gls{padim} \cite{Defard2020PaDiMPatchDistribution} & Frozen       & \boldentry{2.1}{92.1}    & \boldentrynospace{1.1}{1.1}            \\
		 & US \cite{Bergmann2020UninformedstudentsStudent}     & Frozen       & 91.4              & 2.0                      \\ %
		 \midrule
		 & Gaussian fine-tune (ours)                           & Tuned        & 88.7              & 1.7                      \\ %
		 &                                                     & Frozen       & 88.5              & 1.6                      \\
		 & Deep \gls{svdd}                                     & Tuned        & 88.7              & 1.9                      \\ %
		 &                                                     & Frozen       & 88.9              & 2.0                      \\ %
		 & \gls{fcdd}                                          & Tuned        & 89.7              & 1.6                      \\ %
		\bottomrule
	\end{tabular}
\end{table*}
\paragraph{Results.} Assessing results, it can be seen that our proposed fine-tuning scheme improves both \gls{auroc} (\autoref{tab:AS_ROC}) and \gls{pro} (\autoref{tab:AS_pro}) scores.
This is in contrast to Deep \gls{svdd}, where both \gls{auroc} and \gls{pro} decrease upon fine-tuning.
Conversely, the \gls{fcdd} objective also improves \gls{as} performance, and we achieve higher \gls{auroc} values for \gls{fcdd} than reported in \cite{Liznerski2021ExplainableDeepOne}.
Furthermore, differences across the fine-tuning objectives are much lower for \gls{as} than for \gls{ad}. %
Compared to state-of-the-art \gls{as} methods, our method achieves similar \gls{auroc} scores, and slightly lower \gls{pro} scores.

\begin{figure}[tbp]
	\centering
	\subfloat[]{\includegraphics[width=\linewidth]{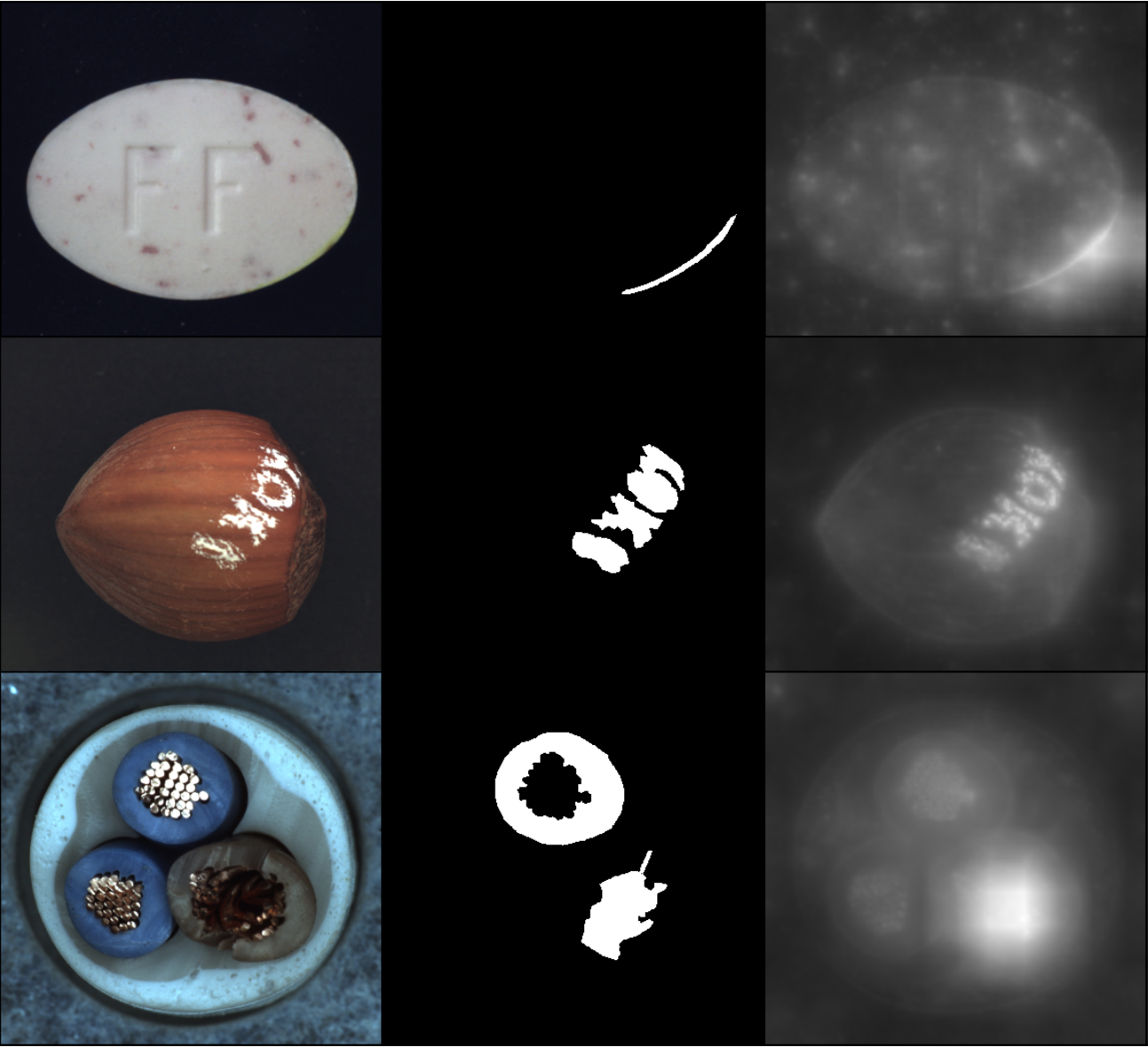}} %
	\vspace{0.5\baselineskip}

	\subfloat[]{\includegraphics[width=\linewidth]{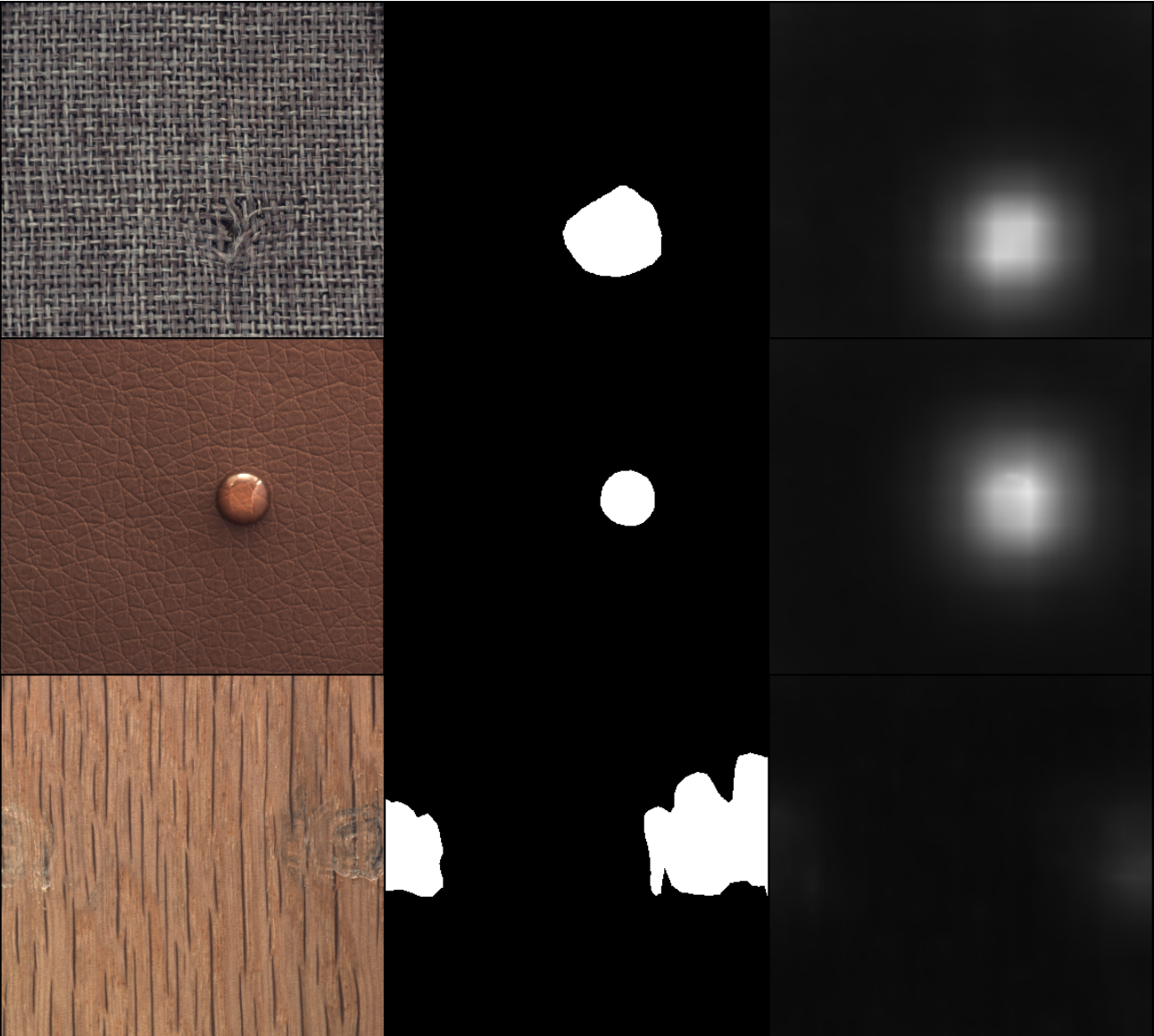}} %

	\caption{Representative successful segmentations as well as failure cases of our approach.
		We show representative successful segmentations as well as failures for three categories (a) as well as three texture (b) classes of the MVTec dataset.
		Left shows the input image, the middle the ground-truth segmentation mask, and the right the heatmap generated by our approach.
		Heatmaps are scaled such that the value range across all test images is mapped to the interval [0, 255].} %
	\label{fig:qualitative_segmentations}
\end{figure}
In addition to quantitative evaluations, we also assess segmentation results qualitatively, showcasing representative results in \autoref{fig:qualitative_segmentations}.
Here, it can be seen that our approach performs well on low-level, textural anomalies (\eg the \enquote{print} on the hazelnut as well as the \enquote{glue} on the leather).
However, it fails to segment/capture high-level, semantic anomalies such as the \enquote{cable swap} or \enquote{scratches} (\enquote{scratches} may be similar in appearance to the background texture in context of the class wood).

\section{\uppercase{Discussion}}
In our work, we have proposed to fine-tune pre-trained feature representations for \gls{ad} using Gaussian distributions, and motivated this by the linkage between generative and discriminative modeling shown in \cite{Lee2018simpleunifiedframework,Kamoi2020WhyisMahalanobis,Rippel2021ModelingDistributionNormal}.
We have also proposed to use augmentations commonly employed for \gls{vrm} to detect onset of catastrophic forgetting, avoiding the bias likely induced by sampling the anomaly distribution for \gls{oe} \cite{Ye2021UnderstandingEffectBias}.

Evaluations on the public MVTec dataset have revealed that the proposed fine-tuning based on the Gaussian assumption achieves a new state-of-the-art in \gls{ad} and improves \gls{as} performance (\autoref{tab:AD_results}, \autoref{tab:AS_ROC} and \autoref{tab:AS_pro}).
Ablation studies demonstrated that onset of catastrophic forgetting can be reliably detected using our proposed early stopping criterion (\autoref{tab:early_stopping_AD}).
We have furthermore shown that the choice of augmentation scheme only has a limited influence on the detection of catastrophic forgetting.
Moreover, ensembling multiple augmentation schemes yields results equal to that of the best scheme applied individually (\autoref{tab:ablation_vicinity}).
Our proposed early stopping regime should thus be transferable also to other fine-tuning approaches and can easily integrate additional augmentation schemes such as the one proposed in \cite{Li2021CutPasteSelfSupervised}.

We furthermore remarked in \autoref{subsec:early_stopping_vrm} that a bias may still be introduced even when using subtle augmentations for model validation rather than \gls{oe}, but argued that this bias would most likely be lower.
Results showed that Deep \gls{svdd} outperforms \gls{fcdd} with respect to \gls{ad} (\autoref{tab:AD_results}), supporting this claim. %
While \gls{fcdd} did outperform Deep \gls{svdd} with respect to \gls{as}, this can be attributed by the segmentation mask generated for the synthetic \textit{Confetti} and \textit{CutOut} augmentations, which provide additional information about anomaly localization leveraged by \gls{fcdd}.
Therefore, one should train \gls{fcdd} without exploiting the synthetic segmentation masks in future work to fully validate our claim.

During our evaluations, we moreover found that \gls{fcdd} achieves better \gls{as} than reported in \cite{Liznerski2021ExplainableDeepOne}.
This can be in part attributed to the fact that we apply \gls{fcdd} to multiple levels of the pre-trained \gls{cnn}.
Furthermore, we use features of an EfficientNet-B4 compared to the VGG16 used in \cite{Liznerski2021ExplainableDeepOne}, and network architectures with better ImageNet performance have been shown to be more suited for transfer learning in \cite{Kornblith2019Dobetterimagenet}.

\subsection{Limitations}
While both \gls{ad} and \gls{as} could be improved by fine-tuning, our approach failed to segment high-level, semantic anomalies (cf.\ \autoref{fig:qualitative_segmentations}).
High-level, semantic anomalies are undoubtedly more difficult to detect than low-level anomalies, as they require a model of normality that is defined on an abstract understanding of the underlying image domain \cite{Ahmed2020DetectingSemanticAnomalies,Deecke2021TransferBasedSemantic}. %
Acquiring such abstract understanding has proven difficult for \glspl{cnn}, as evidenced by their texture-bias \cite{Geirhos2019ImageNettrainedCNNs,Hermann2020OriginsPrevalenceTexture}.
Potentially, learning these concepts could be achieved by employing the self-attention mechanism \cite{Vaswani2017AttentionisAll}, and a lesser texture bias was observed for \glspl{vit} \cite{Dosovitskiy2021ImageisWorth} recently \cite{Naseer2021Intriguingpropertiesvision}.
Thus, one should try to apply features of ImageNet pre-trained \glspl{vit} to the \gls{as} task in future work.
As an alternative, one could also try to leverage features yielded by either object detection or segmentation networks, which have been shown to generate features that maintain stronger spatial acuity \cite{Li2019analysispretraining}.
They have furthermore been shown to improve \gls{as} performance when used as the basis for transfer learning \gls{as} \cite{Rippel2021Leveragingpretrained}.
Moreover, we limited our evaluations to the public MVTec dataset.
In future work, we will therefore revalidate our approach on additional datasets used in literature, such as CIFAR-10 \cite{Ahmed2020DetectingSemanticAnomalies}.
Last, anomalies may also occur for multi-modal normal data distributions \cite{Rippel2021EstimatingProbabilityDensity,Deecke2021TransferBasedSemantic}, and our current fine-tuning procedure can not be applied here.
Here, less constrained priors such as \acrlongpl{gmm} or \acrlong{nf}~\cite{Rezende2015VariationalInferenceNormalizing} may be used instead.

\section{\uppercase{Conclusion}}
In our work, we have demonstrated that fine-tuning of pre-trained feature representations for transfer learning \gls{ad} is possible using a strong Gaussian prior, achieving a new state-of-the-art in \gls{ad} on the public MVTec dataset.
We have further shown that augmentations commonly employed for \gls{vrm} can be used to detect onset of catastrophic forgetting, which typically hinders transfer learning in \gls{ad}.
Here, ablation studies revealed that our method is robust with respect to the chosen synthesis scheme, and that combining multiple schemes is also feasible.
Together, this demonstrated the general applicability of our approach.

\bibliographystyle{./template/scitepress/apalike}
{\small
	\bibliography{./literatur/lit}}
\end{document}